\documentclass[letterpaper, 10 pt, conference]{ieeeconf}  

\IEEEoverridecommandlockouts                              

\usepackage{amsmath,amsfonts}
\usepackage{algorithm2e}
\SetKwComment{Comment}{/* }{ */}
\SetAlFnt{\small}
\SetAlCapFnt{\small}
\SetAlCapNameFnt{\small}
\RestyleAlgo{ruled}
\usepackage{array}
\usepackage{textcomp}
\usepackage{url}
\usepackage{verbatim}
\usepackage{graphicx}
\usepackage{cite}
\usepackage{caption}
\usepackage{subcaption}
\usepackage{svg}
\usepackage{float}
\usepackage{multirow}
\usepackage{booktabs}       
\usepackage{graphbox}
\usepackage{adjustbox}
\let\labelindent\relax 
\usepackage{enumitem}

\begin{document}


\title{\LARGE \bf Control Transformer: Robot Navigation in Unknown Environments through PRM-Guided Return-Conditioned Sequence Modeling}

\author{Daniel Lawson and Ahmed H. Qureshi\thanks{The authors are with the Department of Computer Science, Purdue University, West Lafayette, IN 47907 USA (email: lawson95@purdue.edu; ahqureshi@purdue.edu)}}






\maketitle

\begin{abstract}
Learning long-horizon tasks such as navigation has presented difficult challenges for successfully applying reinforcement learning to robotics. From another perspective, under known environments, sampling-based planning can robustly find collision-free paths in environments without learning. In this work, we propose Control Transformer that models return-conditioned sequences from low-level policies guided by a sampling-based Probabilistic Roadmap (PRM) planner. We demonstrate that our framework can solve long-horizon navigation tasks using only local information. We evaluate our approach on partially-observed maze navigation with MuJoCo robots, including Ant, Point, and Humanoid. We show that Control Transformer can successfully navigate through mazes and transfer to unknown environments. Additionally, we apply our method to a differential drive robot (Turtlebot3) and show zero-shot sim2real transfer under noisy observations.

\end{abstract}


\section{Introduction}

Long-horizon robot control (LRC) from raw local observations is a challenging task requiring a robot to make decisions in a reactive manner, avoiding collisions from partially observed obstacles, while navigating toward the desired goal state. Once solved, it will have widespread applications ranging from search and rescue \cite{liu2013robotic} to autonomous driving \cite{campbell2010autonomous}, where local observations are usually available to the robot system for decision-making. Deep Reinforcement Learning (DRL) \cite{howtoDRLrobo} has demonstrated promise in learning robotic control policies from local observations. However, standard DRL methods struggle with long-horizon problems and are often exhibited in toy tasks with no collision avoidance constraints. 

Recent advancements have led to offline DRL methods\cite{DT,TT} that leverage a robot motion datasets from an expert for training sequence models solving control problems. However, to date, these methods are demonstrated in solving simulated locomotion tasks with either no collision-avoidance constraints or noisy real-world local sensory observations. Furthermore, the assumption of having a large robot skill dataset a priori limits their applicability to scenarios where obtaining robot skills is challenging. The unsupervised robot skill discovery methods \cite{Bagaria2021SkillDF, Sharma2020DynamicsAwareUD} could be used to provide a way to obtain the dataset for offline DRL. However, these methods do not generalize to high-dimensional robot systems such as 23 Degree-Of-Freedom (DOF) Mujoco Humanoid Robot or struggle in exploring complex, cluttered, large environments.

\begin{figure}[t]
  \centering
      \includegraphics[width=\linewidth] {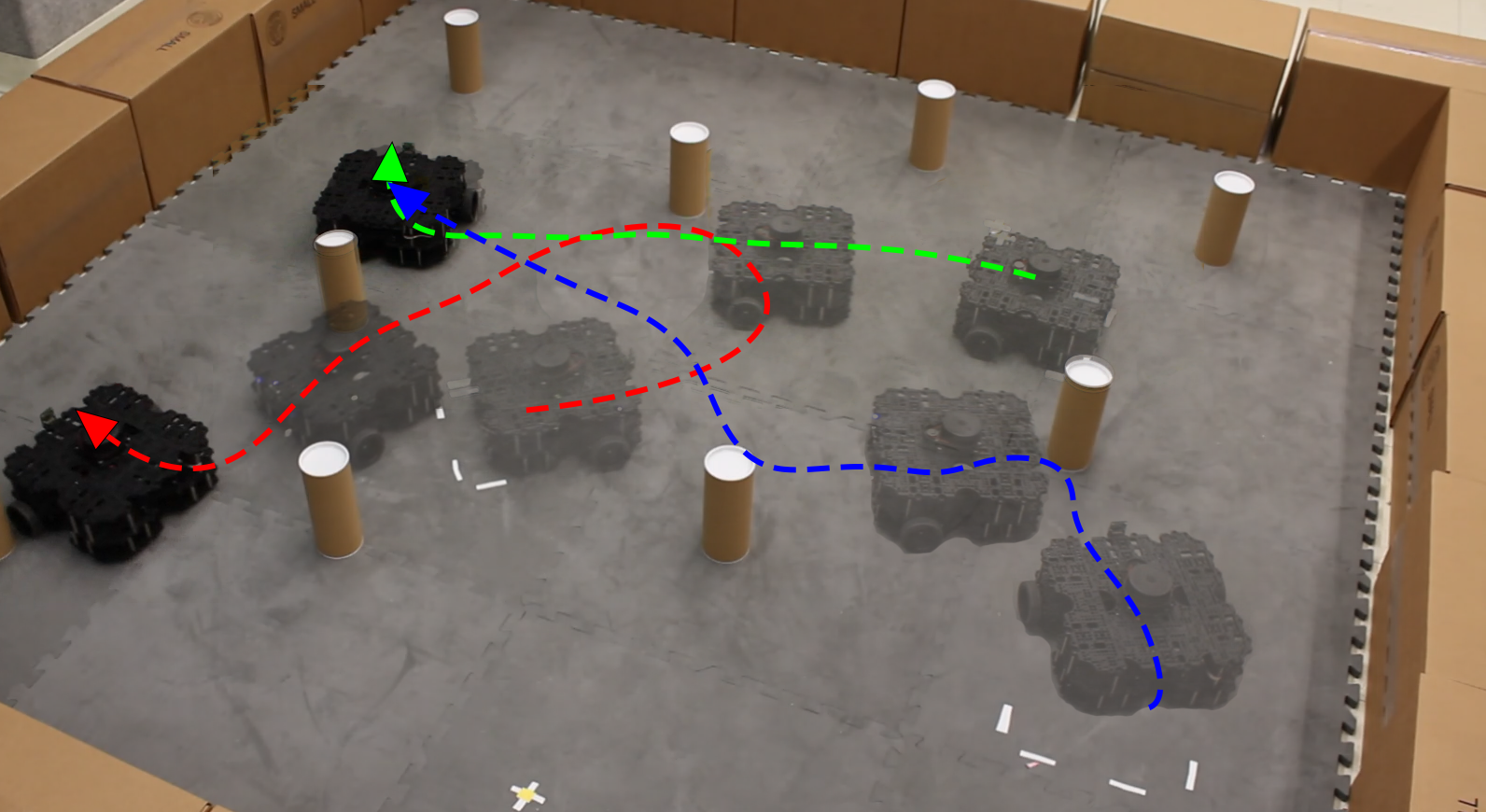}

    
    \caption{Control Transformer navigating in cluttered environments with a differential drive robot, showing sim2real transfer. In each trajectory, the robot is conditioned to reach a target goal position. }
    \label{fig:trajs}\vspace{-0.3in}
\end{figure}

To overcome the limitations mentioned above and scale offline DRL approaches to complex, long-horizon, and real-word robot motion tasks, this paper presents a novel framework called Control Transformer (CT). The CT method solves long-horizon, high-dimensional, robot navigation and locomotion tasks from local raw observations. The main contributions of the paper are summarized as follows:
\begin{itemize}[leftmargin=*]
    \item A scalable Probabilistic Road Map (PRM)-guided method for robot skill discovery that learns robot local policy ensemble over PRM Graph in large, cluttered environments. 
    \item A return-conditioned Transformer framework for sequence modeling of robot behavior using associated local observations. The return conditioning is based on a learnable value function that determines the initial cost-to-go to the target position and guides the Transformer's sequence modeling and generation process.
    \item A fine-tuning strategy that leverages the failures of CT during execution and refines its behavior without catastrophic forgetting, leading to overall better performance in challenging cluttered scenarios.  
    \item Evaluation of the proposed CT method in complex, multi-task simulated and real-world settings with 2 DOF point-mass, 3 DOF differential drive, 12 DOF ant, and 23 DOF humanoid robots. The real-world experiments are with Turtlebot3 demonstrating direct sim2real transfer of our framework. 
    \item The results demonstrate that our method outperforms existing approaches, and to the best of our knowledge, this paper presents the first in-depth analysis and evaluation of the transformer-based offline RL method in solving complex problems and their sim2real transfer with noisy robot states and sensor information (Fig.~\ref{fig:trajs}).
\end{itemize}

\section{Related Work}

\subsection{Hierarchical Reinforcement Learning and Skill Learning}
An approach for extending RL to long-horizon tasks are Hierarchical Reinforcement Learning (HRL) methods \cite{HutsebautBuysse2022HierarchicalRL} which use several policies acting at different levels of abstraction. Many approaches exist, such as sub-goal or feudal HRL methods where higher-level policies provide goals for lower-level policies to execute \cite{Vezhnevets2017FeUdalNF, Nachum2018DataEfficientHR}. Other methods have a higher level policy select lower-level policies, or options, to execute; this can be learned in an end-to-end manner \cite{Bacon2017TheOA} but it can be difficult to transfer options to different tasks. Other HRL methods independently first focus on learning low-level skills in an unsupervised manner, such as empowerment methods, which aim to learn diverse skills that can be used for different downstream tasks \cite{Gregor2017VariationalIC, Eysenbach2019DiversityIA, Sharma2020DynamicsAwareUD}. Deep Skill Graphs (DSG) \cite{Bagaria2021SkillDF} introduces graph-based skill learning using the options framework, and learns a discrete graph representation of the environment. During unsupervised training, a skill graph is incrementally expanded through learning options. For obtaining a low-level policy, we also use graph-based learning but show how we can obtain a single goal-conditioned policy to complete many skills provided by sampling-based planning, rather than many unconditional policies (or options). Rather than having a hierarchy of policies as in HRL, trained in an end-to-end manner, our low-level policy serves to generate data for a more capable transformer policy. 


\subsection{Sampling-based Planning}

Sampling-based planning methods such as Probabilistic Roadmaps (PRM) \cite{PRM} and Rapidly Exploring Random Trees \cite{RRT} are algorithms traditionally used in motion planning to find constraint-free paths from one point to another in a robot's configuration space. This is accomplished through sampling many constraint-free points and connecting neighboring points if they are reachable using local planning. Sampling-based planning has been extended through neural motion planning (NMP) \cite{QureshiMPN, IchterLearningSD, Johnson2021MotionPT} approaches which train neural networks that efficiently plan by learning sampling distributions from expert demonstrations. In contrast, our approach learns by imitating executed actions following sampling-based plans rather than learning the plans themselves as in NMP. Sampling-based planning has been combined with RL in PRM-RL \cite{PRM-RL}, providing a simple method for long-horizon navigation tasks. Sampling-based planning provides robots with high-level plans, where low-level navigation between waypoints in the plan is executed using a learned goal-directed policy. While our method uses PRMs during training, we obtain a policy that can operate without sampling-based planning, allowing it to operate and generalize to environments that are unknown and partially observed. 




\subsection{Conditional Supervised Learning}
Recent work has investigated posing online RL as a supervised learning problem \cite{emmons2021rvs} via goal \cite{Ghosh2021LearningTR} or reward conditioning \cite{Kumar2019RewardConditionedP,Schmidhuber2019ReinforcementLU,trainingupsidown}. Instead of learning a normal policy, which given a state, directly outputs optimal actions, a return-conditioned approach learns to conditionally output actions given both state and desired reward. 
While transformers \cite{Vaswani2017AttentionIA} are an appealing model for sequential decision problems, it has been difficult to adapt transformers to typical online RL \cite{Parisotto2020StabilizingTF}. However, Decision Transformer (DT) \cite{DT}, has found success using transformers by training them as return-conditioned policies. DT operates in the offline setting, requiring high-quality data. 
In this paper, we provide a method for training Decision Transformer style policies without prior expert demonstrations and scaling them to complex problems with collision constraints and noisy real-world observations.

\section{Method}
\begin{figure*}[t]
  \centering
    \includegraphics[width=\linewidth]{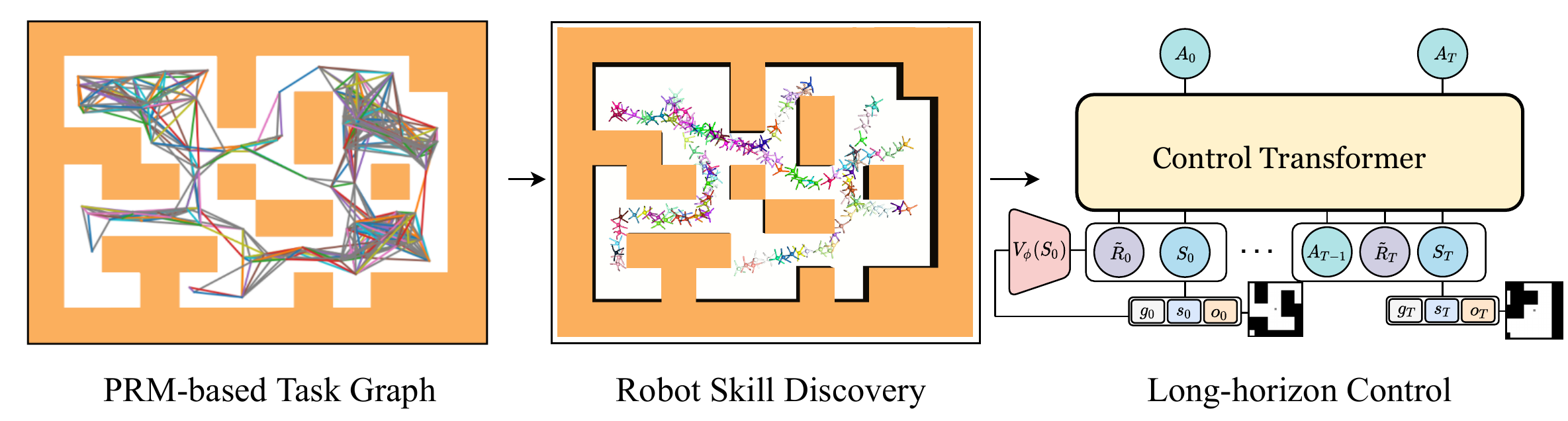}
    \caption{Overview of our learning process. We use PRMs to decompose navigation tasks into discrete graphs, where each edge can be considered a skill. We can use a known low-level controller, or edges can be used as goals for training a a low-level policy with model-free RL. We then guide a controller to complete long-horizon tasks, collecting trajectories. On these trajectories, we train Control Transformer to perform return-conditioned sequence modeling. Afterward, we can optionally fine-tune Control Transformer with planning-guided fine-tuning on failure cases without catastrophic forgetting. }
    \label{mainfig}
\end{figure*}

We now begin to discuss our proposed methods, which are summarized by Algorithm \ref{alg:alg1} and shown in Fig. \ref{mainfig}. 

\begin{algorithm}
\caption{Overall Control Transformer (CT) }\label{alg:alg1}

\KwIn{recovery iterations $\mathit{I}$, trajectory per iteration $\mathit{T}$}
\KwOut{CT $\pi_\theta$, undiscounted value function $V_\phi$}

$\pi^c \gets$ with Algorithm \ref{alg:llc} or set to known low-level controller \\
$\mathcal{T} \gets$ Collect $\mathit{T}$ long-horizon trajectories by guiding $\pi^c$ with sampling-based planning via Algorithm \ref{alg:collection}.  \\
$V_\phi \gets \mathrm{argmin}_\phi ~ \mathbb{E}_{(s_t, \hat{R}_t)\sim\mathcal{T}}[(\hat{R_t} - V_\phi(s_t | g_t))^2]$ via SGD

$\mathcal{L}_\theta = \mathbb{E}_{\tau\sim\mathcal{T}}[ \|a_t - \pi_\theta(a_t | \cdot) \|^2]$ \\
$\pi_\theta \gets \mathrm{argmin}_{\theta}\mathcal{L}_\theta$ via SGD


\Comment{Optional fine-tuning}
\For{$i = 1 \dots \mathit{I}$} {
    Collect $\mathcal{T}^{\mathrm{rec}},\mathcal{T}^{\mathrm{fail}}$  with Algorithm \ref{alg:recovery} \\
    $\mathcal{T} \gets \mathcal{T} ~\bigcup ~\{\mathcal{T}^{\mathrm{rec}},  \mathcal{T}^{\mathrm{fail}}$\}\\
    $\pi_\theta \gets$ continued training with $\mathcal{L}_\theta$

}

\end{algorithm}

\subsection{PRM-based Task Graph}
\label{sec:decomposition}
We first generally discuss how sampling-based planning can decompose long-horizon control problems. We use Probalistic Roadmaps (PRM) \cite{PRM}, obtaining a graph $G = (E,V)$ where vertices are constraint-free points and edges signify that points are reachable using a local planner. We can obtain this graph by sampling $n$ constraint-free points in our environment, and then connect the sampled point to neighbors within connect-distance $d$, forming an edge in the graph, if a constraint-free path exists from one point to another. 


Given a graph $G$ obtained via PRM, and a local controller which we describe as a  policy $\pi^c(a | s, g )$, we can reach any goal point $x_g$ from any start point $x_0$. This is accomplished by finding the closest points in the graph to the start and the goal, $w_0$ and $w_g$, and then using a shortest path algorithm to obtain a sequence of waypoints $(w_0, w_1, \dots, w_g)$. A robot can then traverse from start to goal by taking actions sampled from $\pi^c(a | s, g = w_t - x_t)$, where $w_t$ is the closest waypoint not yet reached at the current timestep, and $x_t$ is the robot position at the current timestep. The sequence of waypoints or plan that guides $\pi^c$ can be fixed, or updated as the robot acts. 

\subsection{Local Skill Learning}
\label{sec:llc}
In this section, we will describe a method for learning $\pi^c$. Optionally, in cases where we have a robot with a simple model and known reliable controller, we could skip this step. We demonstrate the former in our MuJoCo experiments and the latter with a differential drive controller in our Turtlebot3 experiments. We use goal-conditioned RL \cite{Kaelbling1993LearningTA}, modeling our problem through a goal-conditioned Markov decision process (MDP) described by tuple  $(\mathcal{S,G, A},R,p,T,\rho_0, \rho_g)$, where $\mathcal{S,A}$ are the state and action spaces, $\mathcal{G}$ is the goal space, $p$ is the (unknown) transition function $p(s'|s,a)$, $\rho_0$ is a distribution over start states and $\rho_g$ is a distribution over goal states, and $T$ is the maximum length of an episode. In the goal-conditioned setting, the reward, $R$, is also conditioned on the goal, giving $R(s,a | g)$. We aim to find a policy $\pi^*(a|s,g)$ that maximizes expected return, $\mathbb{E}_{s_0 \sim \rho_0, g \sim \rho_g, a_t \sim \pi, s_t \sim p} [\sum_{t=0}^{t=T} R(s_t,a_t | g)]$ \cite{GDRL}. We propose that we can use sampling-based planning to provide goals to train policies. Given a training environment, we first use PRM to obtain a graph $G = (V,E)$. During each episode of training, we sample an edge from the graph which serves as the start and goal for that episode. This process is described in Algorithm \ref{alg:llc}, which is compatible with any goal-directed RL algorithm. Specifically, we use Soft Actor-Critic (SAC) \cite{SAC}  in our experiments with dense rewards that are proportional to progress toward the goal. Low-level policies can be efficiently trained, as the state space of policies only contains proprioceptive information, like robot joint angles and velocities, as they do not need to learn to avoid constraints such as obstacles.


\begin{algorithm}
\caption{Training Low-level Controller}\label{alg:llc}
\KwIn{Graph $\mathcal{G} = (V, E)$,  episodes $\mathit{T}$}
\KwOut{goal-directed policy $\pi^c_\omega(a | s, g)$}
\For{$t = 1 \dots \mathit{T}$}{
    \Comment{Reset environment with new goal edge and robot configuration}
    $x_0, x_g \gets$ edge $e$ sampled uniformly from $E$ \\
    $g_t \gets x_g - x_t$ \\ 
    $\pi_\omega^c \gets$ perform an episode of policy learning
}
\end{algorithm}

\subsection{PRM-guided Skill Discovery}
\label{sec:collection}
Given $\pi^c$ we can use the process defined in Section \ref{sec:decomposition} as a data generating process as described in Algorithm \ref{alg:collection}. This process operates by guiding $\pi^c$, returning a set of trajectories $\mathcal{T} = \{\tau_i\}_{i=1}^{T}$ where $\tau_i = (s^p_0,  g_0, a_t, r_t, \dots s^p_T,  g_T, a_T, r_T)$, which we call planning trajectories. Goals and rewards $g$ and $r$ are set to be with respect to the final goal rather than the waypoints that $\pi^c$ followed. While the original states, $s^p_t$ may only contain low-dimensional proprioceptive information such as robot joint configurations,  we augment the state with high-dimensional exteroceptive observations, $o_{t}$, such as a local map or camera mounted on the robot, providing $s_t = (s^p_t : o_t)$. By adding local observations, a policy trained on this data can learn to act without the need for planning.  Additionally, we can use any reward function $R_\tau(s | g)$ for labeling planning trajectories, which can be different than the reward function used to train a low-level policy. 

\begin{algorithm}[ht!]
\caption{PRM-guided data collection}\label{alg:collection}
\KwIn{number of trajectories to collect $\mathit{T}$, reset interval $\mathit{L}$, Graph $\mathcal{G}$, controller $\pi^c$}
\KwOut{planning trajectories $\mathcal{T}$}
$\mathcal{T} \gets \emptyset$ \\
$\mathrm{successes} \gets 0$ \\
$i \gets 0$ \\
\While{$\mathrm{successes} < \mathit{T}$}{
    \If{$i \mathbin{\%} \mathit{L} = 0$} {
        \Comment{Randomize environment}
        $\mathcal{G} \gets (V, E)$ Sample new PRM over environment, providing robot configurations $V$ and edges $E$.
    }
    $x_0, x_g \gets$ sampled vertices from $V$ \\
    $x_t \gets x_0$ \\
    $\tau^c = (s^p_0, g^c_0, a_0, r^c_t, \dots s^p_T, g^c_T, a_T, r^c_T) \gets$  perform rollout of $\pi^c(\cdot ~| ~s_t, g_t = x_{target} - x_t)$, updating $x_{target}$ with $plan(x_t, x_g, \mathcal{G})$ as waypoints $x_{target}$ reached.  \\
    
    $g_t \gets x_g - x_t$ \\ 
    $r_t \gets R_\tau(s^p_{t+1}| g = x_g - x_t)$ \\
    $s_t \gets (s^p_t : o_t$) \\
    $\tau = (s_0,  g_0, a_t, r_t, \dots s_T,  g_T, a_T, r_T) $ \\
    \If{ $\| x_t - x_g \| < \epsilon$.}{
        $\mathcal{T} \gets \mathcal{T} \bigcup \{\tau\}$ \\
        $\mathrm{successes} \gets \mathrm{successes} + 1$ \\
    }
    $i \gets i + 1$\\
    
}
\end{algorithm}


\subsection{Control Transformer}
This section presents the architecture and training process for Control Transformer. 
\subsubsection{Decision Transformer}
We first begin by explaining the architecture for Decision Transformer \cite{DT}. DT considers RL as a sequence modeling problem, where a sequence is a trajectory $\tau = (\hat{R}_0, s_0, a_0, \dots, \hat{R}_T, s_T, a_T)$ consisting of states, actions, and return-to-go (RTG), $\hat{R_t} = \sum_{i=t}^{T}r_t$. To predict an action, DT uses a deterministic policy $\pi_\theta(a_t | \hat{R}_{h:t}, s_{h:t}, a_{h:t-1})$ where $h=min(1, t-k)$ and $k$ is the context length, which specifies the number of past transitions which are used to predict the next action. After embedding each state, action, and return-to-go, DT uses a GPT-style \cite{Radford2018ImprovingLU} autoregressive transformer decoder to predict actions. During evaluation, we can condition on a target return-to-go for the first timestep which can be equal to the maximum return-to-go achieved in the dataset, or this value times some constant. In some cases, this leads DT to extrapolate, exhibiting behavior that exceeds the best in the dataset. For conditioning during the following timesteps, we can adjust the target return-to-go by the empirical reward, giving $\hat{R}_t = \hat{R}_{t-1} - r_{t-1}$, so one only needs to estimate a good initial value. We refer to an estimate of RTG at time $t$ as $\tilde{R_t}$. We can train a DT by sampling trajectories and minimizing mean-squared error on action predictions, as used in Algorithm \ref{alg:alg1}.

\subsubsection{Control Transformer}
To learn from sampling-guided data collection, we consider approaching our problem as a similar sequence modeling problem, but we consider a goal-directed problem, with trajectories of the form $\tau = (\hat{R}_0, s_0, g_0, a_0, \dots, \hat{R}_T, s_T, g_T, a_T)$, and policy $\pi(a_t | \hat{R}_{h:t}, s_{h:t}, a_{h:t-1}, g_{h:t})$. We also consider the partially-observed multi-environment setting, where a policy may operate in multiple environments with the same underlying task (navigation) but with a different structure for each environment. While we can learn to autoregressively predict actions on this sequence, we would face problems when conditioning our policy, as in DT, the optimal return-to-go is assumed to be constant. This is because we do not know an optimal $\hat{R}_0$ which is dependent on the unknown environment structure, which may change across episodes, and starting state and goal positions.  Thus, we must explore changes that will allow Decision Transformers to be able to generalize to unknown environments, operating from any starting position to any goal. We propose a modification that is compatible with any goal-directed sequential modeling problem.

One approach would be to learn the full distribution of return from offline data, $P_\theta(\hat{R}_t | \hat{R}_{h:t-1}, s_{h:t}, a_{h:t-1})$ \cite{multigamedt}, followed by sampling from this distribution for conditioning during inference time. However, it is difficult to learn the full distribution of return in the goal-directed setting in such a way that we can generalize, predicting return in unseen environments. Instead, we propose learning the mean of this distribution, the undiscounted goal-directed value function $V(s_t | g_t) = \mathbb{E}_{\mathcal{T}}[\sum_{i=t}^{T}r_t | s_t, g_t]$. This is approximated by $V_\phi(s_t | g_t)$, which estimates the expected return-to-go at state $s$ given goal $g$ under the data distribution $\mathcal{T}$. This function is also not conditioned on past history, as at inference time, we predict the first return, $\tilde{R}_0 = V_\phi(s_0 | g_0)$, and use $\tilde{R}_t = \tilde{R}_{t-1} - r_{t-1}$.  We parameterize $V_\phi$ as a separate, neural network and minimize:

$$ \mathcal{L}(\phi) =  \mathbb{E}_{s_t,a_t, \hat{R_t}\sim\mathcal{T}}[(\hat{R_t} - V_\phi(s_t | g_t))^2]$$

To query for more optimal behavior, one could condition on $kV_\phi(s_t | g_t)$, where $k$ is some constant. Alternatively, one can train $V_\phi$ on only the top $X\%$ trajectories, or those that satisfy some condition, such as trajectories where no collision occurred. 

\subsection{Planning-guided Fine-tuning}
A common problem with offline learning relates to distribution shift, where when a trained policy is deployed, the distribution of trajectories encountered while rolling out the policy does not match the training distribution \cite{dagger}. This can cause compounding errors, leading to situations where the policy cannot recover. For example, in navigation, the policy could make slight errors, leading to an unrecoverable obstacle collision. In our framework, the low-level controller guided by planning, may have made few collisions, thus there may be few or no demonstrations of recovering from significant failure in the training distribution. 
To handle this problem, we propose planning-guided fine-tuning as described in Algorithm \ref{alg:recovery}. After offline training, we roll out $\pi_\theta$ until some function $f(s)$ identifies a failure, after which, we take over with $\pi^c$ guided by sampling-based planning. If $\pi^c$ successfully reaches the goal by the end of the episode, then we add two trajectories to our dataset, where the first is an unsuccessful trajectory for reaching the final goal that ends when failure is detected, and the second trajectory demonstrates successfully reaching the goal starting from failure. After augmenting our dataset with recovery and failure trajectories, we can perform additional offline learning.

\begin{algorithm}[h]
\caption{Recovery-based data collection}\label{alg:recovery}
\KwIn{Control Transformer $\pi_\theta$, low-level controller $\pi^c$, value function $V_\phi$, failure identifier $f(s)$, number of recoveries to collect $\mathit{T}$}
\KwOut{trajectories $\mathcal{T}^{\mathrm{rec}}, \mathcal{T}^{\mathrm{fail}}$}

$\mathrm{recoveries} \gets 0$ \\
\While{$\mathrm{recoveries} < \mathit{T}$}{
    $\tau^{\mathrm{fail}} \gets$ rollout $\pi_\theta$ until $f(s)$ returns true, recording start of episode until failure.\\
    $\tau^{\mathrm{rec}}, \mathrm{success} \gets$ take over with $\pi^c$ guided by sampling-based planner as in Algorithm \ref{alg:collection}.\\
    \If{$\mathrm{success}$}{
        $\mathcal{T}^{\mathrm{fail}} \gets \mathcal{T}^{\mathrm{fail}} \bigcup \tau^{\mathrm{fail}}$ \\ 
        $\mathcal{T}^{\mathrm{rec}} \gets \mathcal{T}^{\mathrm{rec}} \bigcup \tau^{\mathrm{rec}}$ \\ 
        $\mathrm{recoveries} \gets \mathrm{recoveries} + 1$ \\

    }
}
\end{algorithm}
\subsection{Architecture details }
 
In this section, we provide additional architectural details for CT.  Obstacle information is represented using local occupancy maps, which are processed by a small convolutional neural network (CNN). Occupancy map representations are then concatenated with both the proprioceptive state and the goal embeddings, resulting in a single state embedding. For a single timestep, embeddings of states, actions, and returns are separately passed as inputs to CT. A shared learned positional embedding is also added to each modality's embedding. When predicting an action, we pass the last $K$ timesteps, feeding $3K$ embeddings to the transformer. With $K > 1$ the transformer has memory over its past interaction. After passing input embeddings through the transformer, we get a transformed embedding for $s_t$ which is passed to a linear layer to predict the next action. The value network similarly processes local occupancy maps with a CNN, as well as additional state and goal information, which is then passed through an MLP which outputs $\tilde{R}_t$. 

\subsection{Hyperparameters and Training Details}
We list hyperparameters for Control Transformer in Table \ref{hyperparams}. For sampling PRMs, we use a connect-distance of $10$ and $200$ sampled points for MuJoCo experiments, and a connect-distance of $2$ and $150$ sampled points for Turtlebot3. For low-level policies trained with SAC, we use networks with 2 hidden layers of size 256 for ant and point as well as 512 for humanoid. During training, we use a batch size of 256 and use a learning rate of $3 \times 10^{-4}$. For rewards in offline data, we use $r(s | g) = -\|g_t\|_2$, with added collision penalties for Turtlebot3. For training low-level policies in MuJoCo, we use distance-based rewards for ant, but find that velocity rewards work better for point and humanoid. We also add an alive bonus for humanoid, which helps prevent the robot from falling over. We implement SAC with Stable Baselines3 \cite{stable-baselines3}, and CQL, a baseline, using \url{https://github.com/young-geng/cql}.
\begin{table}[h]
  \vspace{.025in}
  \caption{Control Transformer hyperparameters used for MuJoCo and Turtlebot experiments.}
  \label{hyperparams}
  \centering
  \begin{tabular}{llllll}
    \toprule
    \shortstack[l]{Parameter} & \shortstack[l]{MuJoCo}  & \shortstack[l]{Turtle} \\
    \midrule
    Number of layers & $4$ & $4$\\
    Number of attention heads & $4$ & $4$ \\
    Embedding dimension & $512$ & $128$ \\
    Batch size & 64 & 128\\
    Nonlinearity function & ReLU & ReLU\\
    Training Context length $K$ & $50$ & $5$ \\
    Evaluation Context length $K$ & $200$ & $5$ \\
    Dropout & $0.1$ & $0.1$\\
    Learning rate & $4\times10^{-4}$ & $10^{-4}$\\
    Weight decay & $10^{-3}$ & $10^{-4}$ \\
    Linear Learning rate warmup & $10^4$ updates & $10^4$ updates\\
    Gradient updates & $7.5 \times 10^4$ & $1.5 \times 10^4$\\ 
    \bottomrule
  \end{tabular}
  
\end{table}
\section{Experiments}
We evaluate our method on simulated MuJoCo robots involving navigation in large mazes and Turtlebot3 navigation in cluttered environments, where we show transfer to a physical robot. We also show that the transformer model allows quick transfer learning to new dynamical systems.


\subsection{MuJoCo experiments}
\label{Mujoco}

We first describe our MuJoCo experiments which are based on the D4RL \cite{d4rl} maze environments, where we modify the observation space to contain local occupancy map observations. The original observation space solely contains proprioceptive  information, where methods must memorize which space is obstacle-free, rather than learn to avoid obstacles. This modification allows us to train and test policies that generalize to unseen maze environments. For our experiments, we test in two maze environments, which include a single training maze (Fig. 3a), that has the structure of ``AntMaze Large'' from D4RL, and an evaluation maze {(Fig. 3b)}, which is unseen during training, testing generalization. This is a difficult task, as we might expect a policy to overfit the structure of the single training environment. In these environments, we test 2 DOF point, 12 DOF ant, and 23 DOF humanoid robots. 

\begin{figure}[H]

  \centering
  \begin{subfigure}[b]{0.49\linewidth}
    \includegraphics[width=\linewidth]{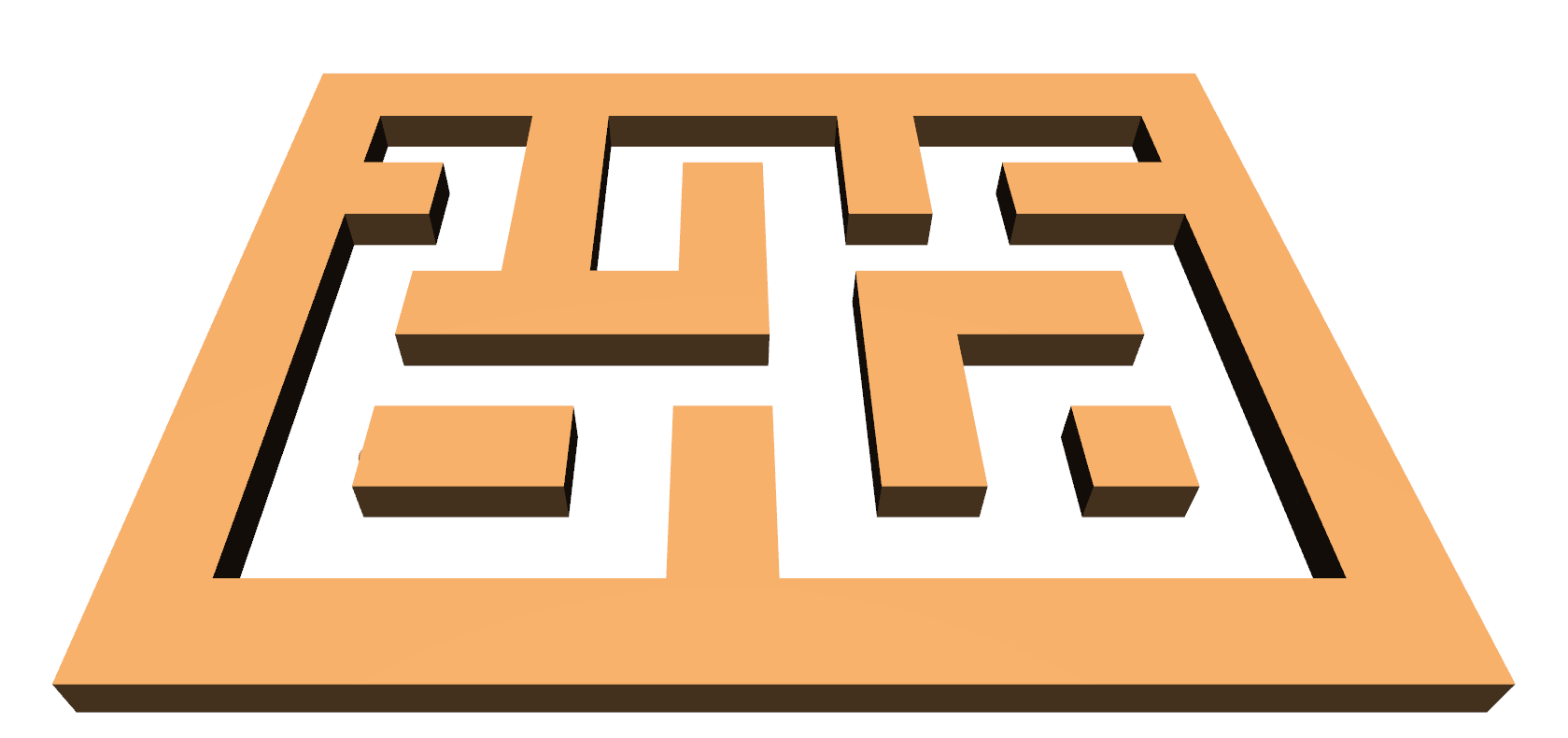}
    \caption{Training maze}
  \end{subfigure}
  \begin{subfigure}[b]{0.49\linewidth}
    \includegraphics[width=\linewidth]{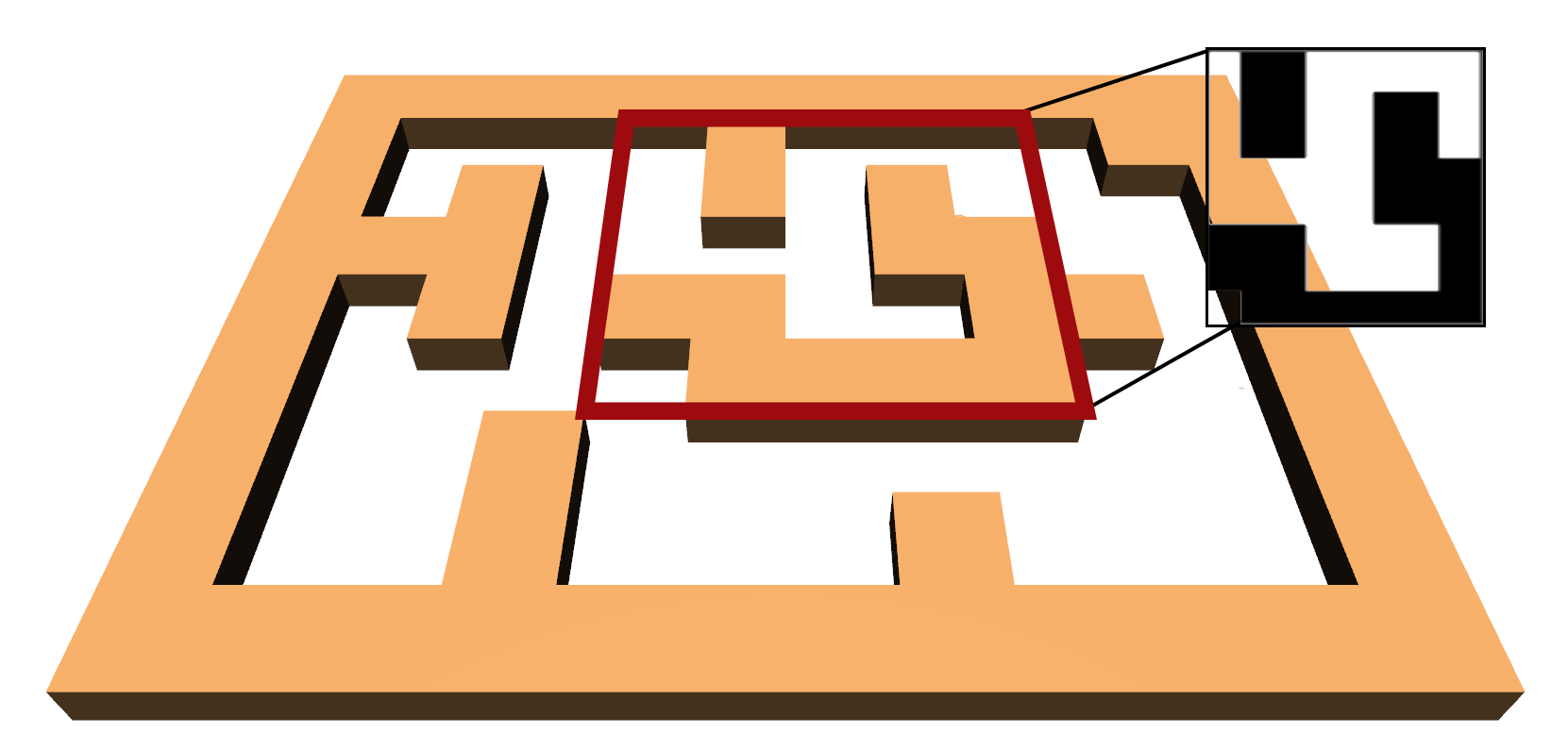}
    \caption{Evaluation maze}
  \end{subfigure}
  \caption{Environments for Section \ref{Mujoco}. In (b), $k\times k$ local map observation where $k=25$ highlighted in red } 
  \label{fig:envs}
\end{figure}

For obstacle information, we use a $2 \times 25 \times 25$ local occupancy map as shown in Fig. \ref{fig:envs}, which has one channel for obstacle information and a second channel that encodes the location of the goal if it is within the local region. For this experiment, we train a low-level controller to obtain skills guided by planning as in Section \ref{sec:llc} with Soft Actor-Critic \cite{SAC}, and then collect trajectories as in Section \ref{sec:collection}, generating 1000 trajectories in the training environment to serve as data for Control Transformer.

\begin{figure}[h]
  \centering
  \vspace{.09in}
  \begin{subfigure}[b]{1.0\linewidth}
    \centering
    \includegraphics[trim={1.5cm 1.1cm 1cm 1.5cm},clip,width=1.00\linewidth]{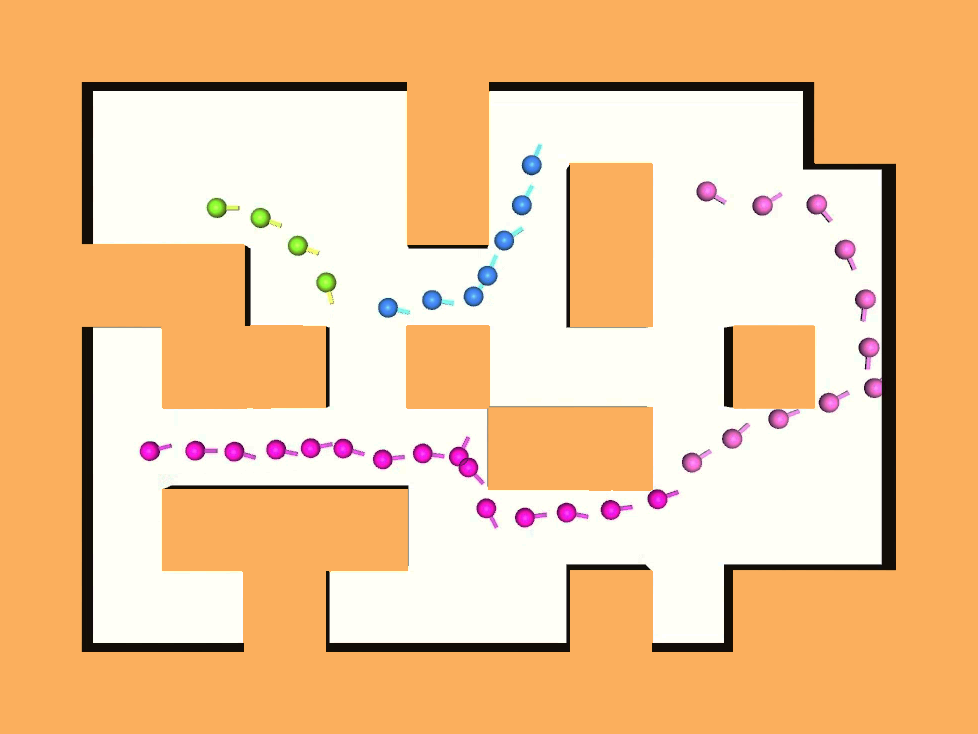}
  \end{subfigure}
  \begin{subfigure}[b]{1.0\linewidth}
    \centering
    \includegraphics[trim={1cm 1cm 1cm 1.3cm},clip, width=\linewidth]{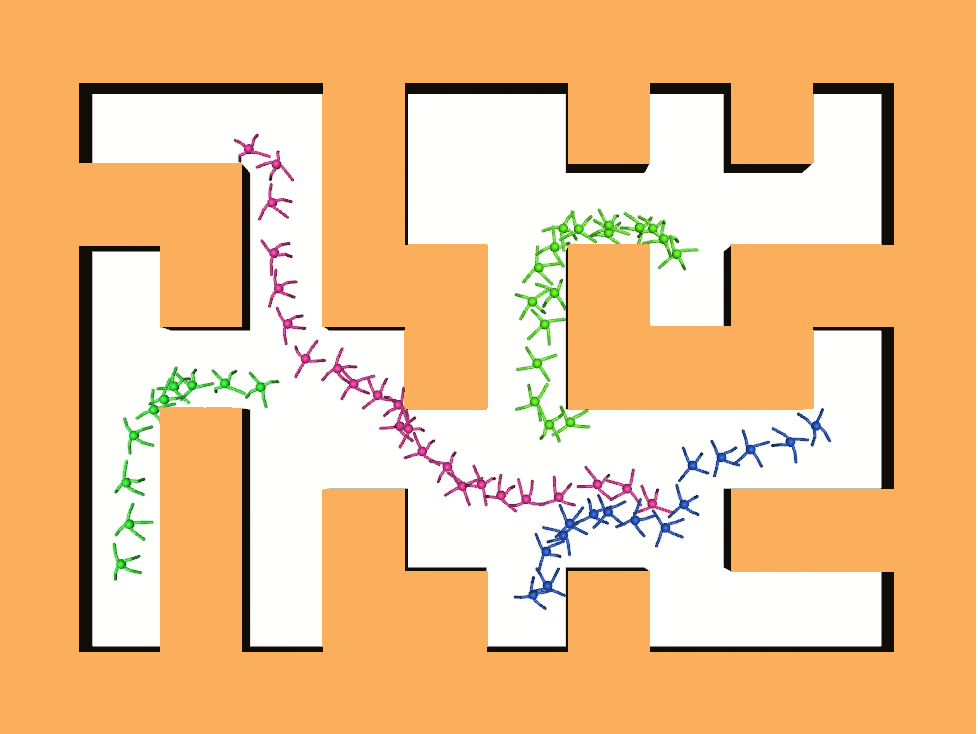}
  \end{subfigure}

  \caption{Control Transformer execution in unseen environments for point, and ant robots.  \vspace{-0.25in}}
  \label{fig:unseen}
\end{figure}

\begin{figure*}[t]
  \centering
    \includegraphics[trim={0.2cm 0.2cm 0.1cm 0.1cm},clip,width=\linewidth]{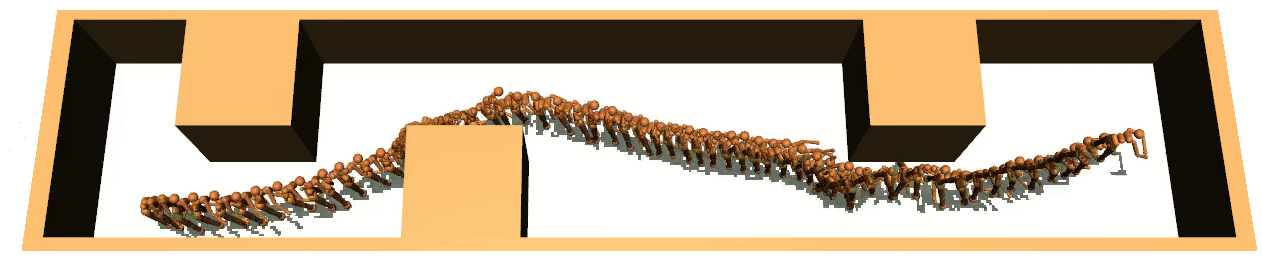}

    \caption{Humanoid trajectory in an unseen hall environment, showing our method's generalization from maze environments.}
    \label{fig:humanoid}\vspace{-0.15in}
\end{figure*}

\begin{table}[h]

  \vspace{5pt}
  \centering
  \caption{Results of Control Transformer (CT), Behavioral Cloning Control Transformer (BC-CT), planning guided fine-tuning Control Transformer (F-CT), and Conservative Q-Learning with planning trajectories obtained with our framework (PT-CQL) on partially-observed maze environments. We report average scaled return per episode followed by average success rate (\%). We average results over 3 trained models for our transformer models. }
  \label{table:experimental-results}
  
  \begin{tabular}{cccccc}
    \toprule
     Env    & Robot &  CT & BC-CT & PT-CQL\\ 
    \midrule
    \multirow{3}{*}{\rotatebox[origin=c]{90}{Seen}}        
    & Point       & -45.67(84.33)      & -48.76(82.67) & -64.46(76)     \\
    & Ant      & -59.28(90)        & -60.90(88) & -83.03(81)  \\
    &  Humanoid     & -222.56(19)        & -243.1(19.67) & -281.77(0.01)  \\
    \midrule 
    \multirow{3}{*}{\rotatebox[origin=c]{90}{Unseen}}  
    & Point         & -120.47(60.33)      & -115.76(57.67) & -132.72(53)      \\
    & Ant          & -89.5(77.33)        &  -86.66(77) &  -112.38(66) \\
    &  Humanoid  & -243.09(18)        & -240.04(19.67)& -308(0.02) \\

    \bottomrule
  \end{tabular}
  \vspace{-8px}
\end{table}

We report our method trained with several variations of our framework trained on the same data. This includes just using planning trajectories, Control Transformer (\textbf{CT}), and training without rewards on planning trajectories, Behavior Cloning Control Transformer (\textbf{BC-CT}). We also compare our approach utilizing return-conditioned sequence modeling to Q-learning on our same collected data with Conservative Q-Learning (CQL) \cite{Kumar2020ConservativeQF}, which is a strong, state-of-the-art method for offline model-free RL using temporal difference (TD) learning. We use CQL, as prior work \cite{DT} finds CQL to be better than other offline TD methods on MuJoCo tasks. We refer to this baseline of CQL trained on our PRM-guided data as planning trajectory CQL (\textbf{PT-CQL}). We also evaluated the state-of-the-art model-free RL method SAC on our test scenarios. However, SAC fails to learn, staying close to $0\%$ success rate; therefore, we exclude it in our comparative analysis. Model-free HRL methods are also excluded as they struggle to solve long-horizon, multi-goal locomotion tasks with collision-avoidance constraints \cite{qureshicomposing}, validating the need for offline RL methods.

We evaluate each offline-RL method on 100 randomly sampled start and goal pairs in both the training and evaluation environment, as shown in Fig. \ref{fig:envs}. We show our results in Table \ref{table:experimental-results}, reporting success and average return for randomly sampled start and goal positions, and find that our transformer-based methods, CT and BC-CT, perform the best. We also show trajectories for ant and point in two additional unseen maze environments in Fig. \ref{fig:unseen}. A trajectory from humanoid in an unseen hall environment Fig. \ref{fig:humanoid} in which our method demonstrates up to a $26\%$ success rate. Note that, in these environments, the success rates of our methods are significantly higher than the baseline approach.  


\subsubsection{Transfer Learning across Different Dynamical Systems}
\label{sec:transfer}
A large amount of work has shown the successes of transformers for transfer learning, where large pre-trained transformers can be adapted for a specific task by fine-tuning on a small amount of data. However, transfer learning has been difficult in  RL. With the adoption of transformers in RL, it may be possible to make more progress. It has been found that initializing Decision Transformers with weights from a language model leads to significantly faster convergence \cite{Reid2022CanWH}. We hypothesize that Control Transformers can be fine-tuned to quickly learn a policy for a different robot.

\begin{figure}[h!]
  \centering
    \includegraphics[width=1\linewidth]{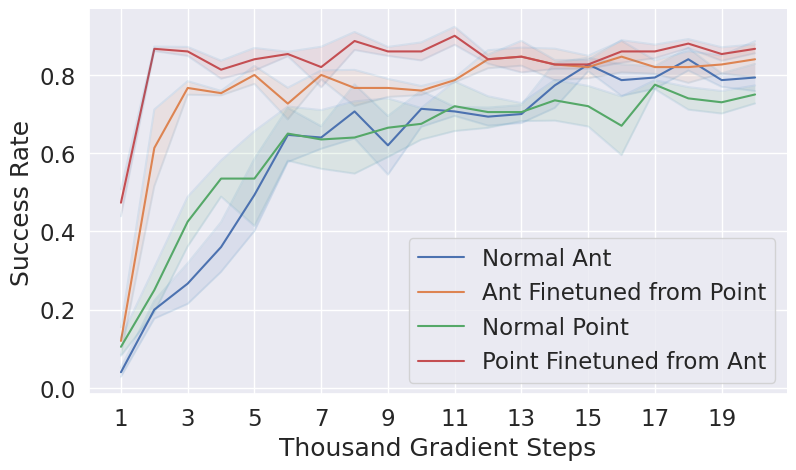} 
    \caption{Control Transformer transfer results. }\vspace{-0.2in}
    \label{fig:convergence}
\end{figure}

To implement this idea, we retain transformer and convolution weights from a trained Control Transformer on a different robot and only re-initialize a few (thousand) weights, particularly the linear embeddings in the input for states and actions, as well as the final layer for action prediction. 
We report our results in Fig. \ref{fig:convergence}, where we show transfer from ant-to-point, and point-to-ant. We find that transfer learning with Control Transformer significantly speeds up training new robots, which may be useful under limited data or computational budgets.   


        

\subsection{ Real World Mobile Navigation}

  
  


  
  

We also evaluate our approach on a Turtlebot3 differential-drive robot. We perform training in randomized cluttered environments simulated with PyBullet \cite{pybullet}, and then transfer Control Transformer to a physical robot, without additional fine-tuning. We use goals $g_t = (\Delta x, \Delta y)$, specifying the translation from the robot's current position to the goal position. We use a proprioceptive state space consisting solely of the difference in yaw between the robot's current orientation and the orientation in the direction of the goal, which is represented by a unit vector $s_p = [\mathrm{cos}(\Delta \psi), \mathrm{sin}(\Delta \psi)]$.  Information about obstacles is provided by a Lidar sensor with a 1-meter radius mounted on the Turtlebot. At each timestep, we project raw Lidar data into a $2 \times 25 \times 25$ occupancy map, allowing us to use the same architecture as in MuJoCo experiments. On the physical robot, we use wheel encoders and IMU to track the robot position, which we must use to calculate  $g_t$ (and $\Delta \psi$).

For this experiment, we demonstrate that our framework can leverage known kinematics models to accelerate training Control Transformer. Instead of learning a low-level controller for data collection, we use a simple controller which outputs linear velocity commands ($V$) proportional to the distance to the goal, and angular velocity ($\omega$) commands proportional to the angle between the robot orientation and the goal. With differential drive robots, the angular velocity for the right and left wheel is defined by $\omega_R = \frac{V + \omega(b/2)}{r}$, $\omega_L = \frac{V - \omega(b/2)}{r}$, where $b$ is the distance between the two wheels and $r$ is the wheel radius. We train Control Transformer in the same manner as in Section \ref{Mujoco}, with an action space of target linear and angular velocities.   
We collect 1000 planning trajectories, and we reset the structure of the environment every 25 episodes, randomly setting new obstacle locations and widths, and sampling a new PRM.


        


        

\begin{figure}[h]
  
  \centering
  
  \begin{subfigure}[b]{.49\linewidth}
    \centering
    \includegraphics[width=\linewidth, align=c]{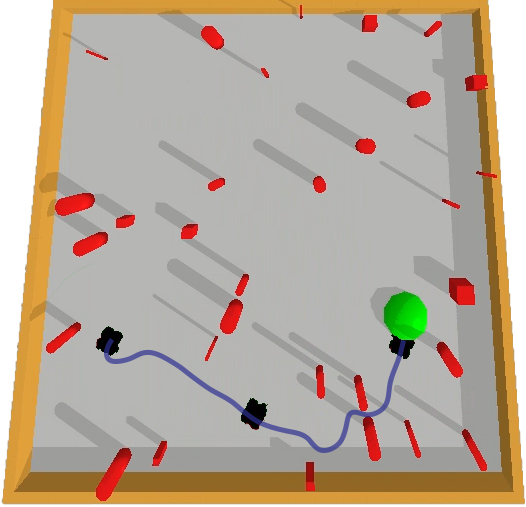}
  \end{subfigure}
  \begin{subfigure}[b]{.49\linewidth}
    \centering
    \includegraphics[width=\linewidth, align=c]{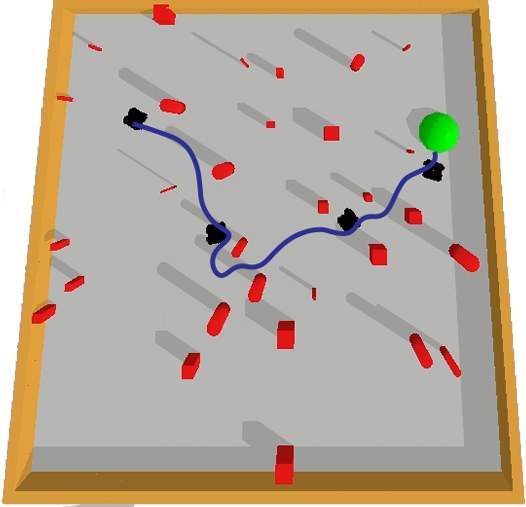}
  \end{subfigure}

   \caption{Trajectories executed on unseen, cluttered environments, with Turtlebot3} \vspace{-0.2in}
   \label{fig:turtlesim}
 \end{figure}

In addition to CT, BC-CT as used in MuJoCo experiments, we also perform further training on recovery and failure trajectories with Algorithm \ref{alg:recovery} with $T = 500$ recoveries and $I = 1$ iterations, which we refer to as planning-guided fine-tuning Control Transformer (F-CT). We compare each method, and report results averaged over $20$ randomized environments and $25$ start-goal pairs per environment. 
We find that BC-CT, CT, and F-CT attain a mean success rate of $87.6 \pm 1.33\%$, $92.47 \pm .94$, and $\mathbf{95.87\pm .9 \%}$ respectively. We find benefits for both return conditioning (CT), as well as additional fine-tuning (F-CT). In Fig. \ref{fig:turtlesim}, we show generalization with environment variation, increasing obstacle density, compared to environments seen during training.


Next, we evaluate zero-shot sim2real transfer, evaluating Control Transformer with planning-guided fine-tuning (F-CT) in the real world. We fix a specific obstacle configuration, as well as 7 start-goal positions. As discussed earlier, Control Transformer is not provided with the structure of the environment, and only uses local LiDAR data projected onto a 2D occupancy map, without any other processing, with a context length of $k = 5$ timesteps. 
We categorize trajectories into three categories, consisting of successful trajectories, where the goal is reached without any obstacle collisions, partially successful trajectories, where the goal is reached, but the robot grazes obstacles, which may occur if the robot makes a turn, but the side of the robot hits an obstacle. A trajectory is a failure if the robot directly runs into the obstacle from the front of the robot, knocks over an obstacle, or continues to push an obstacle as it moves. We report a $5/7 \approx 71.4\%$ partial success rate, and a $3/7 \approx 42.9\%$ full success rate,  showing three trajectories in Fig. \ref{fig:trajs}. Green and red trajectories show two successes, while blue is a partial success. While we find good results in simulation, we believe some shortcomings in the real world are due to inaccurate robot odometry, which is needed for goal-conditioning throughout the duration of an episode, and from noisy LiDAR data.  

\section{Conclusion and Future Work}

In this work, we presented a framework that uses planning with return-conditioned sequence modeling for learning policies capable of long-horizon control tasks such as navigation. We demonstrated that our method learns to navigate on difficult partially-observed mazes and cluttered navigation environments. We also showed that our method generalizes to new environment configurations, and is capable of zero-shot sim2real transfer, without additional real-world data. We also highlight that in one-shot sim2real transfer, the partial failures are primarily due to the robot's noisy odometry and sensor data.

Our future work will introduce state estimation techniques in real robot transfer to overcome noisy observations. In addition, we also aim to map real-world environments using NeRF \cite{mildenhall2021nerf}, performing planning-guided training in environments that more closely resemble reality, using visual observations. Finally, we also plan to expand our work to tasks such as autonomous driving, where it would be possible to also collect human demonstrations, jointly training on a fixed amount of human data, and planning-guided data.  


{
\bibliographystyle{plain}
\bibliography{refs} 
}

\end{document}